\documentclass[a4paper,conference]{IEEEtran}
\usepackage{amsfonts}
\usepackage{url}

\ifCLASSINFOpdf
  \usepackage[pdftex]{graphicx}
  \graphicspath{{../pdf/}{../jpeg/}}
  \DeclareGraphicsExtensions{.pdf,.jpeg,.png}
\else
  \usepackage[dvips]{graphicx}
  \graphicspath{{../eps/}}
  \DeclareGraphicsExtensions{.eps}
\fi
\begin{document}
%
\title{Pyramid Embedded Generative Adversarial Network \\for Automated Font Generation}

\author{\IEEEauthorblockN{Donghui Sun}
\IEEEauthorblockA{Alibaba\\
ruya.sdh@alibaba-inc.com}
\and
\IEEEauthorblockN{Qing Zhang}
\IEEEauthorblockA{Alibaba\\
sensi.zq@alibaba-inc.com}
\and
\IEEEauthorblockN{Jun Yang}
\IEEEauthorblockA{Alibaba\\
muzhuo.yj@alibaba-inc.com}}


%


\maketitle

\begin{abstract}
In this paper, we investigate the Chinese font synthesis problem and propose a Pyramid Embedded Generative Adversarial Network (PEGAN) to automatically generate Chinese character images. The PEGAN 
consists of one generator and one discriminator. The generator is built using one encoder-decoder structure with cascaded refinement connections and mirror skip connections. The cascaded refinement connections embed a multi-scale pyramid of down-sampled original input into the encoder feature maps of different layers, and multi-scale feature maps from the encoder are connected to the  corresponding feature maps in the decoder to make the mirror skip connections. Through combining the generative adversarial loss, pixel-wise loss, category loss and perceptual loss, the generator and discriminator can be trained alternately to synthesize character images. In order to verify the effectiveness of our proposed PEGAN,  we first build one evaluation set, in which the characters are selected according to their stroke number and frequency of use, and then use both qualitative and quantitative metrics to measure the performance of our model comparing with the baseline method. The experimental results demonstrate the effectiveness of our proposed model, it shows the potential to automatically extend small font banks into complete ones.
\end{abstract}


%
\IEEEpeerreviewmaketitle

\section{Introduction}
Font, as one basic element, has been widely used in various aspects of the art and design. However, the design of font is a very time-consuming task. During the first stage of developing a font bank, calligrapher needs to write a large amount of characters as templates . And for the second stage, font designers will spend quite a long time to digitize these character templates and manually create all other characters that do not exist in the templates. Particularly, it takes even more time to design one Chinese font which has a much larger dictionary compared with English or Latin that includes
only tens of letters. Therefore, a more efficient way to automatically generate a font bank is desperately needed.

There are some attempts to synthesize Chinese characters automatically. The most typical method is based on stroke extraction \cite{cao2000model}. In this kind of methods, font generation procedure is divided into stroke extraction and recombination of isolated strokes. Effective stroke extraction from character image plays a crucial role to determine the performance of the font generation. However, the current stroke extraction algorithms  can not always work well due to the complexity and diversity of Chinese characters.

Generative adversarial network (GAN) \cite{Goodfellow2014Generative} is increasingly being used for image generation.  The original GAN has one generator and one discriminator. The generator can generate realistic images and try to fool the discriminator. The discriminator is used to classify real images and generated images. The generator and discriminator can be trained alternately by one minmax policy \cite{Goodfellow2014Generative}. In \cite{mirza2014conditional}, conditional GAN is proposed by feeding an auxiliary information such as image class labels together with original input of GAN to generator and discriminator. In \cite{Isola2016Image}, $pix2pix$ is proposed to transform images from one style to another based on conditional GAN. Similar to $pix2pix$, $zi2zi$ \cite{zi2zi} is also built on conditional GAN and it is the first attempt to automatically generate font images. 

There is another type of method which is not based on adversarial learning and also can be used for image generation. In \cite{Chen2017Photographic}, one convolutional network which is denoted as cascaded refinement network (CRN) is proposed to synthesize photographic images conditioned on pixel-wise semantic layouts. The backbone of CRN is one feedforward convolutional network that consists of several cascaded refinement modules. Each module operates at a given resolution, which is doubled between consecutive modules. 
This design feeds low level information of input image to different layers, preserves much more detailed information during training and results in high-resolution images generation. However, the input and output of CRN are supposed to have identical layout. Thus, it cannot handle the image transformation.



In this work, we treat the task of font generation as image to image transformation and propose a new method for automated font generation. The contributions of this paper are as follows:

\begin{enumerate}
\item 
We propose PEGAN by embedding multi-scale pyramid of refinement information into U-Net \cite{ronneberger2015u}. This design can enable generator preserve much more detailed information and is beneficial for model performance. 
\item We deploy four loss functions in training procedure: adversarial loss, pixel-wise loss, category loss and perceptual loss. The adversarial loss is beneficial to generate images with fine details. The pixel loss indicates the pixel space distance between real and generated images, while perceptual loss measures the discrepancy between them from perceptual aspect. The category loss, which is important for pre-training, enables the model to learn from multiple styles simultaneously. 
\item We build a character set for evaluation based on stroke number and frequency of use. We conduct both qualitative and quantitative measurements to evaluate the performance of proposed font generation model. 
\end{enumerate}

The organization of the rest of paper is as follows. In the next Section, we review recent methods for font generation. In Section \ref{pegan}, the framework of PEGAN is introduced, as well as loss functions and implementation details. Section \ref{exp} shows experimental results, where we report our results using Microsoft HeiTi to generate HuaKang font. Also, we deploy PEGAN performing small font banks extension. Finally, we conclude the paper in Section \ref{conclusion}.

\section{Related Work}
\label{related_work}
In this section we review the relevant work for font generation. There are mainly two categories: stroke extraction based methods and image generation based methods.
\subsection{Font generation based on stroke extraction}
Stroke is the basic unit of Chinese characters. The stroke extraction based methods are the dominant way to generate font in early stage. In \cite{cao2000model}, the strokes of both source font and target font are first extracted. Then autoencoder and self-organizing maps are adopted to cluster the extracted strokes into 100 different groups. Finally, new characters of target font are generated by stroke replacement. In \cite{miyazaki2017automatic}, one automatic extrapolation method for small font is proposed. The strokes of characters from one small subset are first extracted to form a stroke pool. In addition, one transformation matrix between source font and target font is learnt by the corresponding relationship between their skeletons. The missing target characters are generated based on strokes and transformed skeletons. This kind of methods mainly rely on stroke extraction. Moreover, the stroke extraction of calligraphic font is still one challenging problem.
\subsection{Font generation based on GAN}
Recently, GAN is becoming a very hot topic in machine learning community \cite{Goodfellow2014Generative, Chen2016InfoGAN}. As GAN is capable of mapping one distribution into another distribution, these kinds methods can be used to generate realistic fonts.

Inspired by \cite{Isola2016Image}, $zi2zi$ \cite{zi2zi} is the first attempt to generate fonts using GAN. The generator of $zi2zi$ is based on the U-Net structure, as shown in Fig. \ref{fig_unet}. It employs different loss functions, including adversarial loss, L1 loss, and const loss. The $zi2zi$ also introduces category loss such that it can perform one-to-many learning mode. Besides $zi2zi$, there are several GAN based methods for font generation. In \cite{Chang2017Chinese}, the category loss is dropped because it argues that embedded category information can degrade the generation quality for calligraphic fonts. In \cite{Lyu2017Auto}, apart from generator which captures the layout of input font image, it also applys another encoder-decoder network acting as supervisor which reconstructs source images to guide the generator to learn the detailed stroke information.

Compared with stroke based methods, GAN based methods consider font generation as image-to-image transformation. It does not rely stroke extraction and can be trained by end to end. 

\section{Pyramid Embedded GAN}
\label{pegan}
In this section, we introduce the proposed PEGAN for automated font generation. Firstly, we explain the whole framework of the proposed PEGAN. Then, we explain four different loss functions deployed in PEGAN: the adversarial, pixel-wise, category and perceptual losses, respectively. Finally, we illustrate the details of the training procedure. 

\subsection{Framework overview}
We develop PEGAN framework, which consists of an image generator $G$, and a discriminator $D$, as shown in Fig. \ref{fig_pegan}.

\subsubsection{Generator}
\begin{figure*}[!t]
\centering
\includegraphics[width=0.9\linewidth]{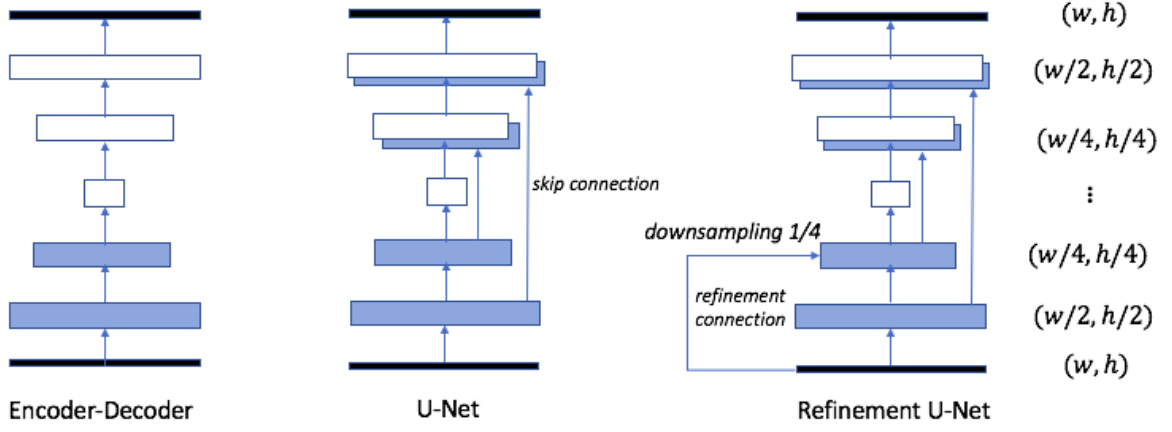}
\caption{Three choices for the architecture of the generator. The common choice is encoder-decoder. The U-Net is an encoder-decoder with skip connections between mirrored layers in the encoder and decoder stacks. We design the refinement U-Net by adding multiple scale inputs for encoder. Besides the original input, downsampled inputs are fed into each module of encoder. The skips from encoder to decoder are also reserved.}
\label{fig_unet}
\end{figure*}

Encoder-decoder network has been used in many previous solutions for image-to-image transformation \cite{hinton2006reducing}. In such a network, the input is  progressively down-sampled and passed through multiple layers until a bottleneck layer, as shown in Fig. \ref{fig_unet}. For image-to-image transformation task the input image of high resolution are mapped to another image. The input and output differ in appearance, but have similar underlying structures. Thus, it should be considered that input and output should share a great deal of low level information.  

In \cite{Isola2016Image}, the structure of U-Net with skip connections is introduced for shuttling this information directly from encoder to decoder. Specifically, skip connections are added between each module $i$ and module $n-i$, where $n$ is the total number of modules. Each skip connection simply concatenates all channels at module $i$ in encoder with those at module $n-i$ in decoder. 

\begin{figure*}[!t]
\centering
\includegraphics[width=0.9\linewidth,height=8cm]{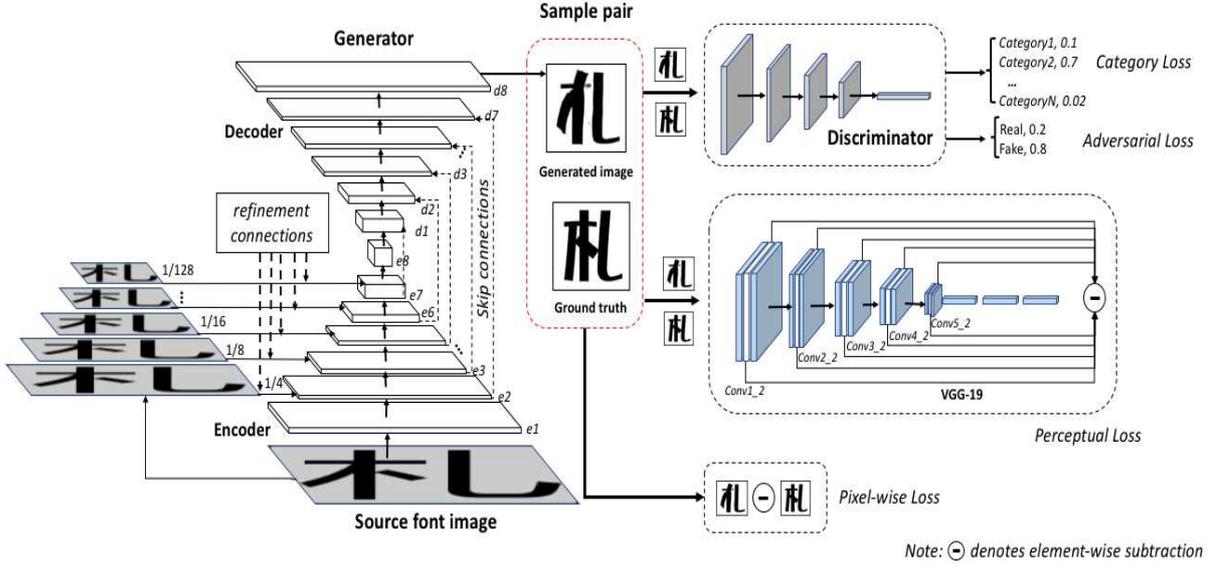}
\caption{The PEGAN framework. The backbone of Generator is U-Net with cascaded refinement connections. $G$ mainly consists of one $Encoder$ and $Decoder$.  Both $Encoder$ and $Decoder$ are formed by a series of convolutional modules, $e_{i}$, $d_{i}, i=1,2,...8$, respectively.  $e_{i}$ is composed of LReLU-Conv-BN. The $e_{1}$ takes source images as input. The rest of $e_{i}$ takes two different inputs: the first are feature maps outputted by preceding modules, the second are downsampled source images. $d_{i}$ is composed by ReLU-DeConv-BN-Dropout. The $d_{1}$ takes $e_{8}$'s feature map as input. Besides the preceding module's feature map, the rest of $d_{i}$ also takes the feature maps generated in $Encoder$. For example, the feature map of $d_{1}$ and $e_{7}$ are concatenated along $depth$ channel and then fed to $d_{2}$. New images generated by $d_{8}$ are fed to discriminator $D$. The $D$ is composed by Conv-LReLU-\{Conv-BN-LReLU\}x3-FC-Sigmoid.  The discriminator $D$ outputs two confidence values, one is for category to which input font images belong, the other one is used to determine whether the input of $D$ are real images or generated images. The pre-trained VGG-19 is used to compute  perceptual loss.}
\label{fig_pegan}
\end{figure*}

Based on this design, we propose a refinement U-Net by embedding a multi-scale pyramid of down-sampled original input into encoder feature maps of different layers. We show our design in Fig. \ref{fig_pegan}. To be more specific, the first module  $e_{1}$ of encoder receives the original source font images ($w\times h$) as input and produces a feature map $F_{1}$ at resolution $\frac{w}{2} \times \frac{h}{2}$ as output. All other modules $e_{i}$ (for i $\neq$ 8) are structurally identical:  $e_{i}$ receives a concatenation of the source font images (downsampled to $\frac{w}{2^{i}}\times \frac{h}{2^{i}}$) and the feature layer $F_{i-1}$ as input, and produces feature layer $F_{i}$ as output. The feature maps, $F_{i}$, are connected to corresponding feature maps in decoder to make mirror skip connections.

\subsubsection{Discriminator}
The discriminator $D$ takes real font images and generated images as inputs, and makes classification between them. $D$ is composed of  Conv and Leaky ReLU (LReLU), followed by three Conv-BN-LReLU module blocks. The last two layers are full connected layer and Sigmoid layer. $D$ outputs two confidence values. One is the predicted probability of which category the input images belongs to and the other is the probability that input images are ``real" images. 
\subsection{Loss functions}
The generator $G$ is trained to produce ``fake'' images that cannot be distinguished from ``real'' images by an adversarially trained discriminator, $D$. In contrary, $D$ is trained to classify ``fake" images from ``real'' images.  Thus, it matters how to measure the discrepancy of ``real'' images and ``fake'' images in training process. Four different loss functions are adopted in PEGAN.
\subsubsection{Adversarial loss}
We begin with the generative adversarial loss from original GAN. The generator $G$ is trained to map samples from noise $p_{z}$  to real distribution $p_{data}$ by one minmax game with discriminator $D$. In training stage, $G$ tries to generate fake images with realistic appearance to confuse $D$. Meanwhile, $D$ aims to distinguish the real samples and the generated samples. 
%
%
As for the task of image to image transformation, we focus on transforming real images in one style to another rather than from random noises to real images, thus the adversarial loss can be written as: 
$$
 \pounds _{adv} = \mathbb {E}_{x \in p_{data}} \left[\log D(x) \right ] + \\
 \mathbb{E}_{z \in p_{input}} \left[\log (1- D(G(z)) \right ]  \eqno{(1)}
$$
where $D$ tries to maximize $\pounds _{adv}$ while $G$ minimizes it.

\subsubsection{Pixel-wise loss}
The most widely used pixel-wise loss for generated images and real images are  L2 distance \cite{dong2016image, shi2016real-time} and L1 distance \cite{Isola2016Image}. It has been found it is beneficial to mix the GAN adversarial loss with traditional pixel wise loss \cite{Isola2016Image}. We use L1 distance rather than L2 as L1 makes less blurring:
$$
\pounds _{L1} =  \mathbb{E}_{x \in p_{data}, z \in p_{input}} \left \|  x - G(z)\right \|_{1}  \eqno{(2)}
$$

\subsubsection{Category loss}
In order to generate high quality font images, it is important to make the model not only aware of its own style, but other font styles as well. Thus it is essential to enable model to learn multiple font styles at the same time \cite{Johnson2016Google}, thereby a category embedding is used by concatenating a non-trainable gaussian noise as style embedding to the character embedding, right before it enters decoder. To prevent the model from mixing the styles together and generating characters that do not look like any of the provided targets, the multi-class category loss is added to supervise the discriminator to predict the style of the generated characters \cite{zi2zi}. In our implementation, the category loss $\pounds _{cate}$ is calculated using sigmoid cross entropy.

\subsubsection{Perceptual loss}
Both the adversarial loss and pixel wise loss are common metrics in image generation task. 
Besides, we use another loss function that is closer to perceptual similarity. The perceptual loss explores the discrepancy between high-dimensional representations of images extracted from a well-trained CNN \cite{johnson2016perceptual}. We define the perceptual loss based on ReLU activation layers of the pre-trained VGG-19 network described in \cite{simonyan2015very}. Layers of different depths can represent image features at different abstraction levels: from edges, colors to object patterns. Matching both lower-level and higher-level activations in the perception network guides the synthesis network to learn  fine-grained details as well as global arrangements \cite{Chen2017Photographic}.

Let $\phi$ denotes the layers of VGG-19, for one training pair $\left \langle x,z \right \rangle$ ( $x \in p_{data}, z \in p_{input}$ ), the perceptual loss is defined as

$$
\pounds _{perp} = \sum_{l} \lambda_{l} \left \|  \phi_{l}(x) - \phi_{l}(G(z))\right \|_{1} \eqno{(3)}
$$
We use the configuration in \cite{Chen2017Photographic}. For layers $\phi_{l} $ $(l \geq 1)$, the `$conv1\_2$', `$conv2\_2$', `$conv3\_2$', `$conv4\_2$', and `$conv5\_2$' in VGG-19 are chosen to calculate perceptual loss. The $\lambda$ is weight of different layers. In our implementation, we set $\lambda_{1}=\lambda_{2}=\lambda_{3}=\lambda_{4}=1, \lambda_{5}=10$.

In total, the loss of PEGAN is weighted sum of these four losses.
$$
\pounds _{PEGAN}=  w_{adv} \pounds _{adv} +  w_{L1}\pounds _{L1} + w_{perp}\pounds _{perp} + w_{cate}\pounds _{cate} \eqno{(4)}
$$ 
Here we set $w_{adv}$ = $w_{perp}$ = $w_{cate}$ = 1, $w_{L1}$ = 100.
\subsection{Implementation details}
In this paper, all images are scaled into 256$\times$256.  We take data augmentation by firstly enlarge images and then random cropping.  Apart from VGG-19, all convolutional and deconvolutional layers in the above mentioned parts have 5$\times$5 kernel size and 2$\times$2 stride. We choose initial learning rate of 1e-3 and train the proposed model with Adam \cite{kingma2015adam} optimization method.

The entire training process consists of \textit{pre-training} stage and \textit{tuning} stage. In the \textit{pre-training} stage, one source font is used to generate $N$ different fonts. We denote this as one-to-many \textit{pre-training} mode. In the \textit{tuning} stage, we choose one font from these $N$ fonts as the final target. In addition, we freeze the encoder of generator $G$ and update the rest parameters during tuning. In our implementation, we set $N$=20.

\section{Experiments}
\label{exp}
In this section, we first explain the experimental configuration and the measurements used for image quality evaluation. Then, we evaluate the proposed PEGAN and compare our result with that of baseline. 

\subsection{Evaluation setting up and Metrics}
It is an open problem that image quality evaluation in image generation task. The quality evaluation of generated character images are different from other image generation tasks. The basic element of Chinese character is stroke. In general, characters with more strokes are much more difficult to generate with high quality, thus different characters have different difficulty to create. Considering this, we build one evaluation character set according to the number of strokes and frequency of use. As shown in Table \ref{table_easy_mid_hard}, we define three levels of difficulty based on stroke numbers.
Each level contains 100 most commonly used characters.

\begin{table}[htp]
\caption{The evaluation character set}
\begin{center}
\begin{tabular}{c|c|c|c}
\hline
Level  & Easy & Mid & Hard \\
\hline
Stroke number& $\leq$ 5 & $\geq$ 6 $\&$ $\leq$ 9 & $\geq$ 10 \\
\hline
 Character number & 100 & 100 & 100 \\
 \hline
\end{tabular}
\end{center}
\label{table_easy_mid_hard}
\end{table}%

In this work, we used qualitative and quantitative metrics to evaluate the performance of proposed font generated model. For qualitative evaluation, we deployed the `Visual Turing Test' introduced in \cite{shrivastava2017learning}, and designed a simple user study where subjects were asked to classify images as real or generated. 
We had 14 subjects in this visual test. Each subject was shown a random selection of 20 real character images and 20 generated character images in a random order, and was asked to label the character images as either real or generated. We use average classification accuracy as qualitative metrics. Meanwhile, we chose several commonly used image quality methods for quantitative evaluation, such as Peak Signal to Noise Ratio (PSNR), Structural Similarity Index (SSIM) \cite{wang2004image}, Universal Quality Index (UQI) \cite{Wang2002A}. Given one pair images, PSNR estimates absolute errors, while SSIM and UQI are perception-based models that consider image quality degradation as perceived change in structural information, and they also incorporates important perceptual phenomena, including both luminance masking and contrast masking terms \cite{wang2003multiscale}. 

Besides above two kinds of measurements, we also applied one evaluation method based on character recognition model. We assumed one moderate character recognition model trained using a lot of real images could recognize the generated character images correctly if generated characters have basic shape and layout. On the other hand, if the generated character images suffer from serious quality problem such as lacking of strokes and stroke incompleteness, they cannot be recognized by the model. Thus, the recognition accuracy can be used to indicate the quality of the generated character images.

\subsection{Microsoft HeiTi generates HuaKang POP3 Std W12}
\begin{figure}[!t]
\centering
\includegraphics[width=0.9\linewidth]{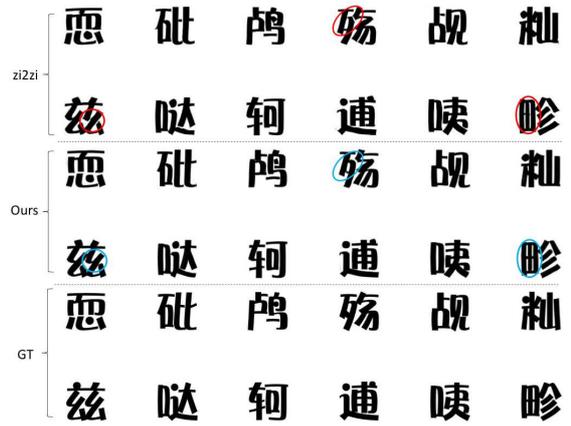}
\caption{The comparison between generated KuaHang POP3 Std W12 using the proposed PEGAN and $zi2zi$ \cite{zi2zi}, ground truth (GT). In general, $zi2zi$ also has a good result. However, incomplete strokes exist in some characters. We use red circles label these bad cases in $zi2zi$`s result. These bad cases get improved by the proposed PEGAN.}

\label{fig_huakang}
\end{figure}

This font, HuaKang POP3 Std W12, is very popular for calligraphic designers. We tried to use Micsosoft HeiTi to generate this font. In \textit{pretraining} stage, we sampled 2000 characters from Microsoft HeiTi and another 20 target fonts as training data, and compared our method with baseline $zi2zi$ method which is built on conditional GAN. The generated result is shown in Fig. \ref{fig_huakang}, from which we come up with a conclusion that some incomplete strokes that emerges in $zi2zi$ are improved a lot by the proposed PEGAN. 

The comparison of quantitative evaluation result are shown in Table \ref{table_eval}. Our proposed PEGAN outperforms $zi2zi$ at first three metrics, means that PEGAN can generate images with less absolute errors of pixel content and perceptual structure information. In term of character recognition accuracy, the generated images by PEGAN is  easier to be recognized correctly than $zi2zi$. The result of Visual Turing Test was shown in Table \ref{table_turing}. The average human classification accuracy of PEGAN output is 65\%, while that of $zi2zi$ result is 73\%. The lower human classification accuracy means that  character images are generated with more realistic appearance so that they can confuse human. It can be concluded that the proposed PEGAN improve the quality of generated character images.

\begin{table}[htp]
\caption{The comparison for quantitative evaluation results of PEGAN and baseline}
\begin{center}
\begin{tabular}{c|c|c|c|c}
\hline
& PSNR & SSIM & UQI & Character recognition model \\
\hline
$zi2zi$ \cite{zi2zi} & 17.45 & 0.83 & 0.26 & 95$\%$ \\
\hline
PEGAN & 17.52 & 0.84 & 0.27  & 96$\%$ \\
\hline
\end{tabular}
\end{center}
\label{table_eval}
\end{table}%

\begin{table}[htp]
\caption{The comparison of Visual Turing Test Result}
\begin{center}
\begin{tabular}{c|c}
\hline
& Classification Accuracy \\
\hline
$zi2zi$ & 73\% \\
\hline 
PEGAN & 65\% \\
\hline
\end{tabular}
\end{center}
\label{table_turing}
\end{table}%

\subsection{Small font banks expansion}
\begin{figure}[!t]
\centering
\includegraphics[width=0.85\linewidth]{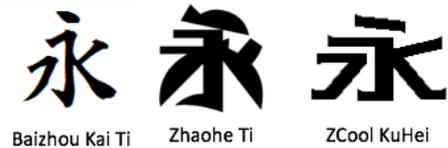}

\caption{Style demonstration of three small fonts.}
\label{fig_smallfonts}
\end{figure}

\begin{figure}[!t]
\centering
\includegraphics[width=0.85\linewidth]{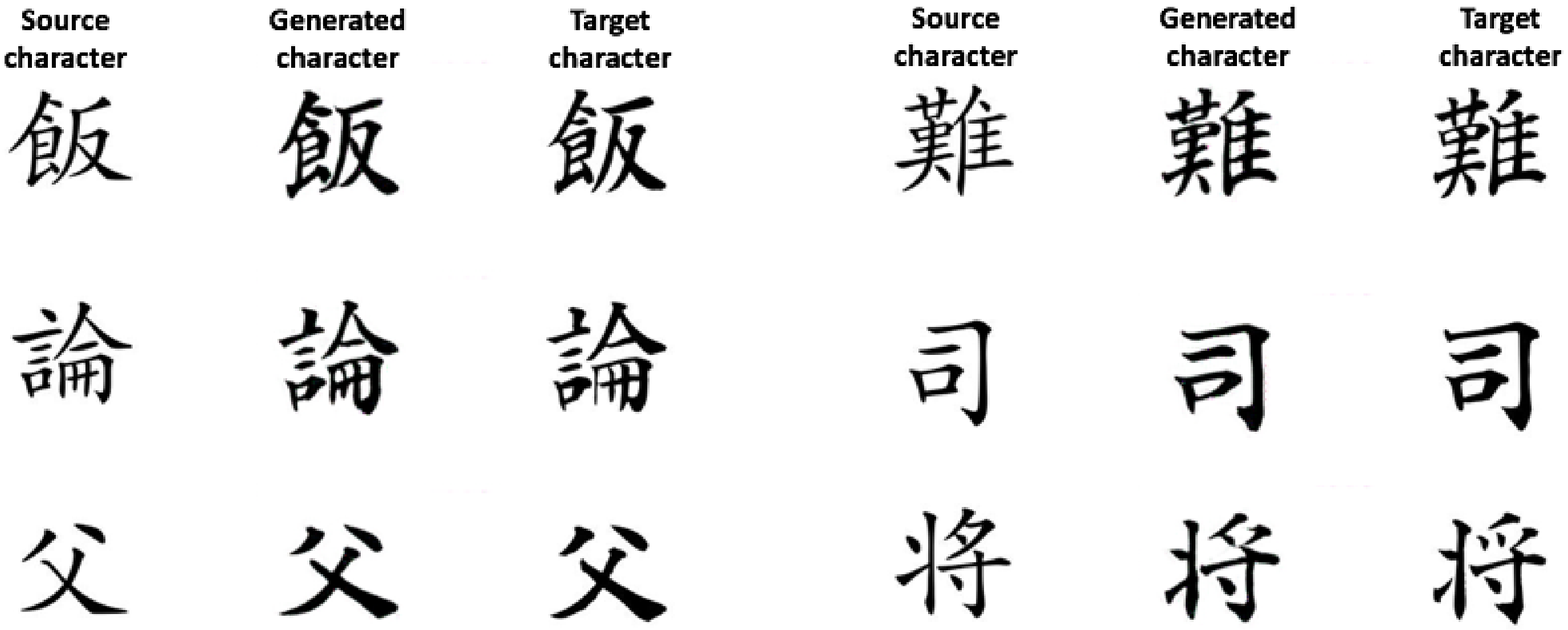}
\caption{The generated Baizhou KaiTi images (testing data) using PEGAN. From left to right: Source character, Generated character, Target character ground truth. The number of training sample for Baizhou KaiTi is very limited. It only contains 670 Chinese characters.  We also achieve satisfactory results by using the proposed PEGAN.}
\label{fig_baizhou}
\end{figure}

\begin{figure}[!t]
\centering
\includegraphics[width=0.9\linewidth]{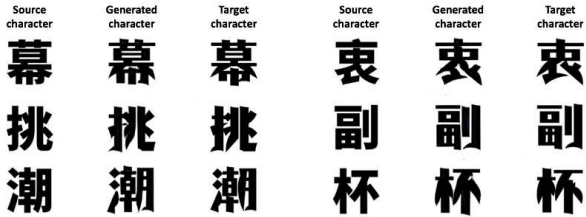}
\caption{The generated ZhaoHeTi testing data using PEGAN.}
\label{fig_zhaohe}
\end{figure}

\begin{figure}[!t]
\centering
\includegraphics[width=0.88\linewidth]{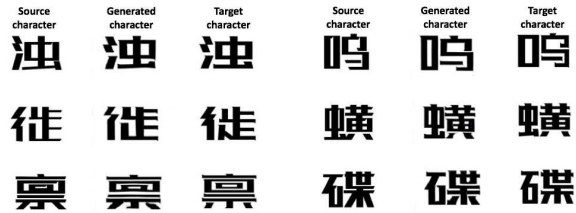}
\caption{The generated ZCool KuHei testing data using PEGAN,}
\label{fig_zcool}
\end{figure}

Besides the generation of HuaKang POP3 Std W12, we apply PEGAN expanding three small font banks: Baizhou KaiTi, ZhaoHeTi and ZCool KuHei. The number of Chinese character in this three fonts are relatively small but all of them are often used in the design of package and banner. We show some examples of these three font styles in Fig. \ref{fig_smallfonts}. Baizhou KaiTi is one Japanese style font containing about 670 Chinese characters. ZhaoHeTi is a Japanese style art font with about 2000 Chinese characters. ZCool KuHei is a crowdfunding font, and it contains 3563 Chinese characters. The three generated characters are shown in Fig. \ref{fig_baizhou}, Fig. \ref{fig_zhaohe} and Fig. \ref{fig_zcool}. All the characters shown in the figures are not included in the training data. We can find that even these characters are not seen by the model during training process, it can use the knowledge obtained by training to handle new characters and achieve style transforming successfully. As a result, the new proposed PEGAN is capable of generating supplementary characters for a font style and automatically extend small font banks into complete ones (6763 characters).

\section{Conclusion}
\label{conclusion}
In this paper, we propose Pyramid Embedded GAN to synthesize Chinese characters automatically. In PEGAN, the generator mainly consists of U-Net structure with cascaded refinement connections and skip connections. We embed a multi-scale pyramid of down-sampled original input into encoder feature maps of different layers on the one hand, on the other hand, we also shuttle the multi-scale feature maps from encoder to corresponding feature maps in decoder by using skip connections. 

In order to measure the discrepancy of real images and generated images in training stage,  four different loss functions are deployed: adversarial loss, category loss, pixel-wise loss and perceptual loss. We split the training process into $pre-training$ stage and $tuning$ stage. For a proper evaluation of generated character image quality, we build one evaluation character set based on stroke number and frequency of use. Moreover, we take both qualitative and quantitative metrics to measure the generated character image quality, and build one evaluation method based on character recognition model as well. To verify the performance of our proposed PEGAN, we perform experiment using Microsoft HeiTi to generate HuaKang POP3 Std W12, both qualitative and quantitative evaluation result indicates the effectiveness of PEGAN. We also use PEGAN to extend three small font banks. Although the minimum number of Chinese character of this small fonts is only 670, we still obtain a good result to make more characters for this font. In the future, we will investigate new methods to further improve the fine detailed information of generated character image and try to develop font bank for business use.

\bibliographystyle{IEEEtran}
\bibliography{refs}

\end{document}